\documentclass[11pt,a4paper]{article}
\usepackage[hyperref]{eacl2021}
\usepackage{times}
\usepackage{latexsym}

\usepackage{microtype}

\usepackage{graphicx}
\usepackage[tone]{tipa}
\usepackage{multirow}
\usepackage{siunitx}
\usepackage{threeparttable}
\usepackage{url}
\usepackage[shortlabels]{enumitem}
\usepackage{arydshln}
\usepackage{caption}
\usepackage{subcaption}
\usepackage{float}

\aclfinalcopy 

\author{Yevgen Matusevych \\
  School of Informatics \\
  University of Edinburgh \\
  \texttt{yevgen.matusevych@ed.ac.uk} \\\And
  Herman Kamper \\
  E\&E Engineering\\
  Stellenbosch University \\
  \texttt{kamperh@sun.ac.za} \\ \AND
  Thomas Schatz\\
  Department of Linguistics \& UMIACS\\ University of Maryland \\
  \texttt{tschatz@umd.edu} \\\And
  Naomi H. Feldman\\
  Department of Linguistics \& UMIACS\\ University of Maryland \\
  \texttt{nhf@umd.edu} \\\AND
  Sharon Goldwater \\
  School of Informatics\\
  University of Edinburgh \\
  \texttt{sgwater@inf.ed.ac.uk} \\}

\date{}

\title{A phonetic model of non-native spoken word processing}

\begin{document}
\maketitle

\begin{abstract}
Non-native speakers show difficulties with spoken word processing. Many studies attribute these difficulties to imprecise phonological encoding of words in the lexical memory. We test an alternative hypothesis: that some of these difficulties can arise from the non-native speakers' phonetic perception. We train a computational model of phonetic learning, which has no access to phonology, on either one or two languages. We first show that the model exhibits predictable behaviors on phone-level and word-level discrimination tasks. We then test the model on a spoken word processing task, showing that phonology may not be necessary to explain some of the word processing effects observed in non-native speakers. We run an additional analysis of the model's lexical representation space, showing that the two training languages are not fully separated in that space, similarly to the languages of a bilingual human speaker.
\end{abstract}

\section{Introduction}

Compared to native speakers, non-native speakers perform differently in a variety of tasks related to auditory language processing, both at the phone and at the word level. At the phone level, these tasks usually require speakers to compare individual phones (e.g., phone discrimination or identification), while spoken word processing tasks usually test the implicit activation of a certain word in the memory (e.g., lexical priming, word translation). In some cases, non-native speakers' behavior is consistent across the tasks: lower performance in spoken word processing tasks is directly associated with difficult phone contrasts. For example, upon hearing a word \textit{rock}, Japanese speakers activate both \textit{rock} and \textit{lock} in their lexical memory \cite{cutler2004}, probably because they find it difficult to discriminate the English \textipa{[\*r]}--\textipa{[l]} phone contrast \cite{miyawaki1975}. In other cases, however, non-native speakers' behavior in spoken word processing tasks cannot be explained by difficult phone contrasts \citep{cook2016, amengual2016, darcy2012}.
For example, in a translation task native English speakers may confuse Russian words \textit{moloko} \textipa{[m@\textltilde2"ko]} (`milk') and \textit{molotok} \textipa{[m@\textltilde2"tok]} (`hammer'), even though this pair of words does not have a difficult phone contrast \citep{cook2016}.

This dissociation between the behavior in phone discrimination vs.\ spoken word processing tasks has been attributed to different kinds of representations involved. On the one hand, thanks to phonetic knowledge, speakers recognize individual phones in a given language. On the other hand, speakers store phonological representations of the words they know in their mental lexicon and use those representations to recognize spoken words \cite[e.g.,][]{pallier2001}. 
It is often implicitly assumed that any lexical processing effect should be attributed to stored phonological representations of the words \citep{gor2020, cook2016, cook2015, darcy2013, darcy2012, mcqueen2006}. At the same time, \textit{phonetic} effects are not limited to the perception of individual phones: some existing theories argue that phonetic details are encoded in the lexical memory \citep[e.g.,][]{pierrehumbert2002, hawkins2003, port2007}. This raises the question: can some of the spoken word processing effects normally attributed to phonology be instead explained in terms of phonetic perception?

In this study, we use computational modeling to test a hypothesis that some spoken word processing effects observed in non-native speakers can be explained without involving phonolexical representations---by phonetic perception, which results from phonetic learning, or speakers' attunement to the sounds of their native language \citep{werker1984}.
We use a model developed for speech technology applications \citep{kamper2019}. Earlier, it was used to simulate early phonetic learning and successfully predicted some infant phone discrimination data \citep{matusevych2020b}. This model learns from natural speech data that is not segmented at the phone level. It never receives information about individual phones in isolation or about phone-level differences between words. Therefore, the model is not equipped with an explicit mechanism to learn abstract phonolexical representations, making it a good candidate to test our hypothesis: if our model without the knowledge of phonology can correctly predict a particular effect, that effect can at least partially be attributed to phonetic learning. Because our goal is to see whether phonetic perception can explain some of the existing data in principle, even just one phonetic model making correct predictions about the data would be a positive result.

By design, lexical processing tasks require at least minimal knowledge of the target language. That is why they are normally carried out with bilingual speakers \citep[or second language learners:][etc.]{gor2020, amengual2016, cook2016}. In bilingual speakers, the two languages interact at various levels, including lexical \citep[e.g.,][]{weber2004, sunderman2006}. To take this into account, we simulate bilingual speakers by training the model on two languages simultaneously.

We present three simulations. The first two show that the model exhibits predictable behaviors in discrimination tasks. In the third one, we present a case study to test whether a lexical processing effect commonly attributed to phonological representations can be explained in terms of phonetic learning alone, without the influence of phonology.
In addition, we examine whether the representations in our bilingual model match the pattern observed in bilingual lexical access.
Existing studies \citep[][etc.]{weber2004, lagrou2011, shook2012} show that upon the presentation of a word, competitor words in both languages may get activated (non-selective lexical access). We carry out a language classification task with our model, showing that the two languages are not fully separated in its representation space.

\section{Method}

\subsection{Simulations}

We train a computational model---a correspondence autoencoder recurrent neural network \citep[CAE-RNN;][]{kamper2019}---on speech data from one or two languages in order to simulate monolingual and bilingual speakers. 
We then test these different versions of the model on discrimination tasks and compare the observed patterns to those found in human speakers. This general methodological framework is adopted from \citet{schatz2019}.

We run three simulations, described in more detail in the respective sections below. Although it would be ideal to use the same set of languages in each simulation, the choice of languages is limited by available results from studies with human participants. We use language pairs for which human data is available and where our model has previously been tested on the target languages in the monolingual context.
In Simulation~1, we look at phone discrimination by infants exposed to two languages (English and Mandarin), showing that the model correctly predicts a discrimination pattern observed in such infants \citep{kuhl2003}.
In Simulation 2, we show that the model can predict discrimination effects at the word level, observed in adult native English speakers and Japanese learners of English \cite{mackain1981}.
In Simulation~3, we show that the result obtained in a translation judgment task with native Russian speakers and English learners of Russian \citep{cook2016} can be at least partially explained in terms of phonetic learning, without the effects of phonology.

\subsection{Model}

\begin{figure}
	\centering
	{\includegraphics[scale=0.25]{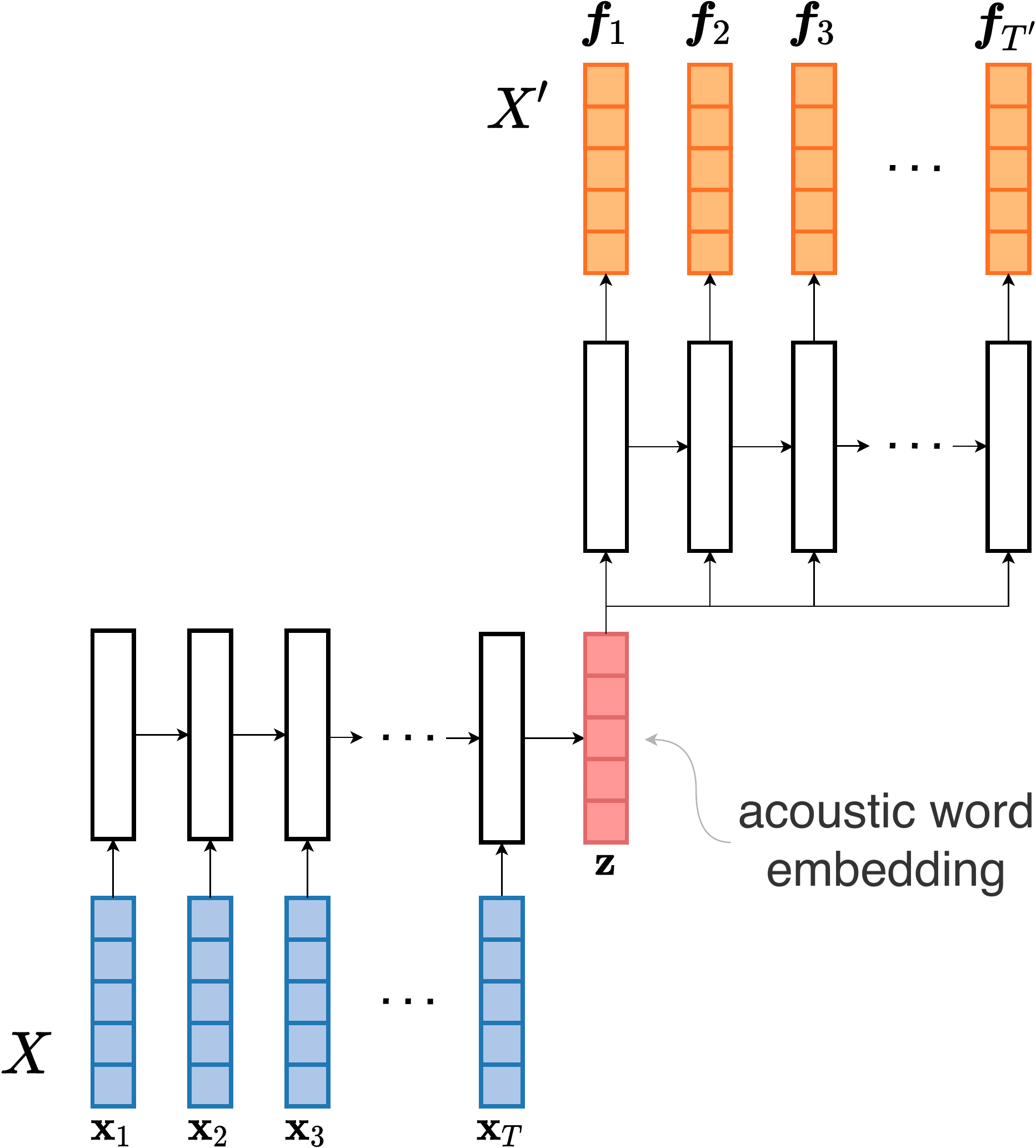}}
	\caption{The model learns to reconstruct an acoustic instance of a word, $X^\prime$, from another acoustic instance of the same word, $X$.}
	\label{fig:model}
\end{figure}

The CAE-RNN \citep[][]{kamper2019} is an extension of a recurrent autoencoder \citep{chung2016}, in which both encoder and decoder are recurrent neural networks. Unlike an autoencoder, the CAE-RNN is trained on pairs of word tokens of the same type (e.g., two acoustic instances of the word \textit{apple}). It receives one instance of a word (represented as a speech sequence), encodes it into a vector of a fixed dimensionality (an \textit{acoustic embedding}), and then tries to reconstruct the other instance in the pair, as shown in Figure~\ref{fig:model}.
Formally, each training item is a pair of acoustic words $(X, X')$. Each word is represented as a sequence of vectors: {$X = \left( \vec{x}_1, \ldots, \vec{x}_{T} \right)$ } and {$X' = \left( \vec{x}'_1, \ldots, \vec{x}'_{T'} \right)$}. The loss for a single training item is:
\begin{equation}
\ell(X, X') = \sum_{t = 1}^{T'} ||\vec{x}'_t - \vec{f}_t(X)||^2
\end{equation}
where $X$ is the input and $X'$ the target output sequence, and $f_t(X)$ is the $t^{\textrm{th}}$ decoder output conditioned on the embedding $\mathbf{z}$.
At inference time, we can encode a sequence of arbitrary duration (e.g., a phone or a word) into a fixed-dimensional acoustic embedding in the model's representation space.

We choose this model because it showed promise for the study of human cognition: it correctly predicted some patterns of infant phonetic learning \citep{matusevych2020b}, and some of its basic properties are compatible with human auditory cognition and lexical access \citep{matusevych2020a}. The advantage of this model compared to others \citep[e.g.,][]{schatz2019} is its ability to represent speech sequences of any duration in a common representation space (the embeddings have a fixed number of dimensions), in which perceptual similarity between sequences can be computed using a simple distance function. The model handles individual phones and acoustic words in exactly the same way, allowing us to easily generalize from phone-level to word-level representations. In addition, the model has been successfully trained on multiple languages for a speech technology application \citep{kamper2020b, kamper2020}, potentially making it a good candidate for simulating bilingual speakers.

Following earlier studies \citep{kamper2019, matusevych2020b}, we first pretrain the model as an autoencoder RNN for $15$ epochs without early stopping using the Adam optimization \citep{kingma2015} with a learning rate of $0.001$. We then train the model for $3$ epochs on $100\textrm{k}$ ground truth pairs from either one or two languages as described next. We use $3$ hidden layers ($400$ gated recurrent units each) in both the decoder and the encoder, and an embedding dimensionality of $130$.

\subsection{Training and test data}

\begin{table}
\centering
\begin{threeparttable}
\begin{center}
A. Training data.
\end{center}
\begin{tabular}{p{0.5cm}p{1.2cm}p{1.7cm}>{\raggedleft\arraybackslash}p{1.1cm}>{\raggedleft\arraybackslash}p{1.1cm}}
\hline
\multicolumn{1}{p{0.5cm}}{Sim.\ \#} & Language         & Corpus        & Duration (hh:mm)     & No.\ of spk.\\ \hline
\multirow{4}{*}{1}                & EN  & WSJ\tnote{1}                   & 19:30 & 96 \\
                                  & JA & GlobalPhone\tnote{2} &  19:33    & 96 \\
                                  \cline{2-5}
                                  & EN  & Buckeye\tnote{3}        & 9:13     & 20\\
                                  & JA & CSJ\tnote{4}         &  9:11     & 20 \\ \hline
\multirow{4}{*}{2}                & ZH & AIShell\tnote{5}     & 58:59    & 166 \\
                                  & EN  & WSJ                   & 58:49    & 166 \\
                                  \cline{2-5}
                                  & ZH & GlobalPhone               & 11:51    & 48 \\
                                  & EN  & WSJ                     & 11:49    & 48\\ \hline
\multirow{2}{*}{3}                & RU                   & GlobalPhone  & 11:07     & 58\\
                                  & EN                   & WSJ    & 11:07    & 58\\ \hline                                  
\end{tabular}

\begin{center}
B. Test data.
\end{center}
\begin{tabular}{p{0.5cm}p{1.2cm}p{1.7cm}>{\raggedleft\arraybackslash}p{1.1cm}>{\raggedleft\arraybackslash}p{1.1cm}}
\hline
\multirow{2}{*}{1}                & \multirow{2}{*}{EN}  & WSJ                     & 9:39     & 47\\
                                  &                           & Buckeye      & 9:01     & 20 \\ \hline
\multirow{2}{*}{2}                & \multirow{2}{*}{ZH} & AIShell              & 58:45    & 165 \\
                                  &                           & GlobalPhone          & 11:51    & 48 \\ \hline
\multirow{1}{*}{3}                                 & \multirow{1}{*}{RU}                   & GlobalPhone   & 11:01     & 57 \\
\hline
\end{tabular}
\begin{tablenotes}
           {\footnotesize \item[1] Wall Street Journal CSR corpus \citep{paul1992}.}
           {\footnotesize \item[2] Multilingual text and speech database \citep{schultz2002}.}
           {\footnotesize \item[3] Buckeye corpus of conversational speech \citep{pitt2005}.}
           {\footnotesize \item[4] Corpus of spontaneous Japanese \citep{maekawa2003}.}
           {\footnotesize \item[5] Open-source Mandarin speech corpus \citep{bu2017}.}
     \end{tablenotes}
\end{threeparttable}

\caption{Corpus samples used in the simulations.}
\label{tbl:corpora}

\end{table}

The model is trained on isolated words and tested on either phones or words extracted from corpora of natural speech based on existing forced alignments \citep[][]{matusevych2020b, kamper2020}. All speech data is encoded using a common approach in speech processing: each speech sequence is divided into $25$-ms-long frames (sampled every $10$ ms), from which $13$ Mel-frequency cepstral coefficients (MFCCs) are extracted using Kaldi \citep{povey2011}. 

The subsets of the corpora that we use are listed in Table~\ref{tbl:corpora}. Within each pair in part A of the table, the subsets are matched on the number of speakers, their gender, and the amount of data per speaker. This ensures that we only compare models trained on the same amount and type of data. In Simulations 1 and 2, we follow the setup from a previous study and use two subsets from different corpora per language. In Simulation 3, we could not obtain two different corpora of Russian speech, and instead train the model five times with different random initializations on the same data.

In case of bilingual models, we train them simultaneously on both languages, using mixed input. We use the relative amount of training data in each language as a simple proxy variable for language proficiency: the higher the model's relative exposure to a language, the higher its proficiency in that language. In bilingual training, we use the same total amount of data as for the corresponding monolingual models---in terms of both the number of tokens (for pretraining the model) and of training pairs. For example, consider a monolingual English model and a monolingual Mandarin model, each trained on $10\textrm{k}$ tokens and $100\textrm{k}$ pairs. Then for training a `balanced' bilingual model we take the $5\textrm{k}$ most frequent tokens from English and Mandarin each, generate $50\textrm{k}$ pairs in each language, and use the combined $10\textrm{k}$ English--Mandarin token data for pretraining and the combined $100\textrm{k}$ pairs for training the CAE-RNN.

\subsection{Simulating discrimination tasks}

To test a model's ability to discriminate a particular phonetic or lexical contrast, we use the machine ABX task \citep{schatz2013}, which is standard in research on zero-resource speech technology for evaluating discriminability of speech units \citep[][etc.]{versteegh2015, dunbar2017, dunbar2020} and is commonly used for simulating human speech discrimination tasks \citep[e.g.,][]{martin2015, schatz2019, millet2019}. The machine ABX task allows us to easily design precise comparisons (e.g., compare words that only differ in $1$ phone) and is not sensitive to the absolute distances in the embedding spaces, which may vary across simulations.

In the ABX task, A and X are two instances of the same word type (e.g., \textit{right}), while B is a different word type (e.g., \textit{light}). If A and X are closer to each other in a model's representation space than B and X, the model's prediction is correct, otherwise it is not. Irrespective of the test units (phones, words), an acoustic segment in our model is represented by a single vector. We perform the ABX task directly on the vectors, allowing us to compare segments of different duration (without doing any type of alignment). Following earlier studies, we use angular cosine distance to measure the distance between the stimuli in the embedding space. The model is evaluated by considering the proportion of ABX triplets for which it makes correct predictions: $0\%$ error corresponds to perfect discrimination, and $50\%$ to chance performance. 

To test whether the difference between the ABX error rates of several models is significant, we fit mixed-effects regressions to the error rates of these models. Significance for the  effect of interest is then determined using two-tailed ANOVA tests (with Satterthwaite degrees of freedom approximation) on the predicted values of the regressions.

\section{Simulation 1: Phone discrimination}

Previously, the monolingual CAE-RNN was shown to correctly predict the crosslinguistic difference in the discrimination of the Mandarin \textipa{[C]}--\textipa{[\texttctclig \super h]} contrast \citep{matusevych2020b}, observed in Mandarin-learning vs.\ English-learning infants \citep{tsao2006}. Considering this, it may seem trivial to show that a model with some exposure to Mandarin data (bilingual) would also achieve lower error than a model with no such exposure (English monolingual). However, potentially complex interactions between training languages may result in an unpredictable phonetic space. This is why we need to ensure the bilingual model behaves as expected, before we move on to the word-level tasks. To do this, we run a simple sanity check: whether a model trained on two languages behaves in a predictable way on the same Mandarin contrast. 

\subsection{Setup}

We focus on the experiment of \citet{kuhl2003}, who showed that exposing English-learning infants to a small amount of Mandarin Chinese improves their ability to discriminate \textipa{[C]}--\textipa{[\texttctclig \super h]}. Our goal is to test whether the model also correctly predicts the pattern for English-learning infants with vs.\ without exposure to Mandarin.

It is difficult to estimate how the infants' amount of Mandarin exposure in the experiment maps onto the English--Mandarin ratio of training data in our model.  We therefore try the ratios of $90$:$10$ and $75$:$25$ (to simulate an infant with a higher exposure to English than to Mandarin), as well as $50$:$50$ (a control condition, a balanced bilingual). As a baseline to compare our bilingual models to, we train the model on English speech alone ($100$:$0$ ratio). For reference, we also train a model on Mandarin speech alone ($0$:$100$).
Using each model, we embed a set of \textipa{[C]} and \textipa{[\texttctclig \super h]} phones from the test corpus and run a \textipa{[C]}--\textipa{[\texttctclig \super h]} discrimination task.
We expect each bilingual model to show lower error than the English monolingual model.

\subsection{Results}

\begin{figure}
  \centering
      \includegraphics[width=\linewidth]{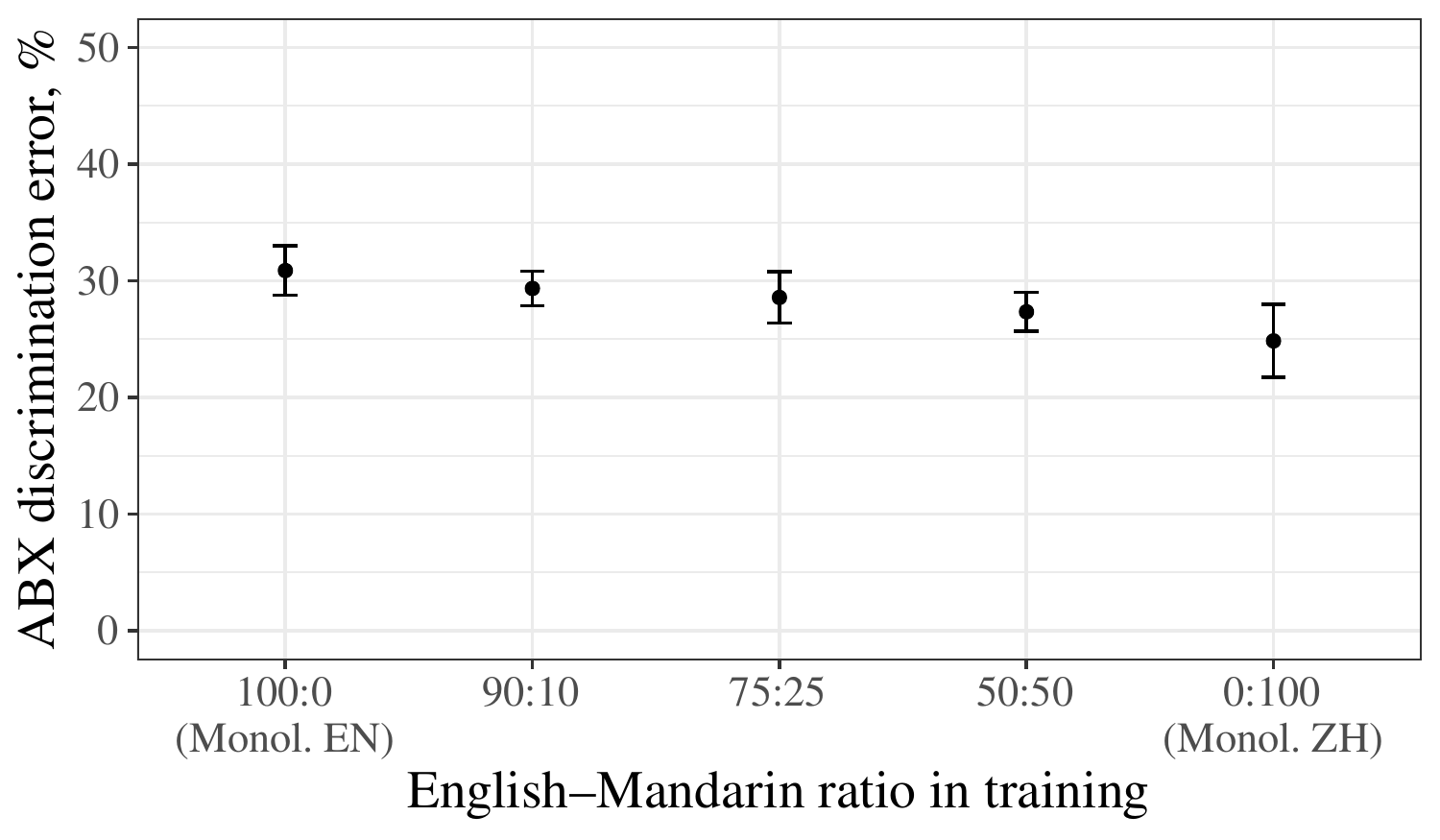}
      \caption{Models' ABX error rates in the Mandarin \textipa{[C]}--\textipa{[\texttctclig \super h]} phone discrimination task. Error bars show standard error of the mean over two training corpus samples $\times$ two test samples.}
  \label{fig:mnen}
\end{figure}

The ABX error rates of the models (see Figure~\ref{fig:mnen}) show the expected pattern: the higher the exposure to Mandarin, the lower the error rates in the target discrimination task. Even having $10\%$ of Mandarin data ($90$:$10$ model) on average results in $1.6\%$ reduction in absolute error compared to the monolingual English model. A mixed-effects regression fitted to the models' error rates shows that this difference is not significant, but the other two bilingual models do show significantly lower error rate than the monolingual English model: we observe a $2.2\%$ error reduction in the $75$:$25$ model and $3.5\%$ in the $50$:$50$ model compared to the English baseline.
This suggests that even a relatively small amount of training data in a given language (under $25\%$) can improve the model's ability to discriminate between some contrasts in that language, consistent with the empirical findings of \citet{kuhl2003} with infants. 
To summarize, the bilingual CAE-RNN model behaves as we expected: it can correctly predict infant-like behavior in phone discrimination.
In the next simulation, we test the model on a word discrimination task.

\section{Simulation 2: Word discrimination}

Simulation 2 tests whether our bilingual model behaves in predictable ways at the word level. Recall that it represents a sequence of any given duration as a fixed-dimensional vector. The compression of a dynamic speech sequence that unfolds in time into a `static' vector results in information loss.
Since words are normally longer speech sequences than phones, it is not obvious whether the model's behavior in word discrimination and phone discrimination will be consistent. To examine this, we test the model on minimal pairs of words with [\textipa{\*r}]--\textipa{[l]}, a phone contrast on which the monolingual CAE-RNN previously showed an infant-like crosslinguistic discrimination pattern \citep{matusevych2020b}. As before, we train the model on one or two languages and see if it behaves in a predictable way, this time in a \textit{word} discrimination task.

\subsection{Setup}

\citet{mackain1981} tested adult native speakers of American English and native Japanese learners of English on the discrimination of English words \textit{rock--lock} (i.e., [\textipa{\*r}]--\textipa{[l]} contrast). Learners with low English proficiency  scored nearly at chance in this task, while highly proficient learners showed the discrimination scores close to those of native English speakers. This result is also in line with studies showing that native Japanese speakers' discrimination of [\textipa{\*r}]--\textipa{[l]} can improve after relevant phonetic training in English under certain conditions \citep[e.g.,][]{strange1984, logan1991, bradlow1997, iverson2005}. Our goal is to test whether our model can correctly predict this word-level discrimination pattern.

Similarly to Simulation 1, we train the model on Japanese (for reference) or English speech alone (as the baseline to compare to), or on a combination of the two in proportion $90$:$10$, $75$:$25$, or $50$:$50$, to simulate native Japanese learners of English with variable proficiency. 
For each model, we embed a set of acoustic words ([\textipa{\*r}]--\textipa{[l]} minimal pairs) from the test corpus.
As in the original experiment, we compare the bilingual models to the native English model: to make correct predictions, a bilingual model needs to show higher ABX discrimination error than the English model. Also, we expect the error rate to decrease with higher exposure to English: $90$:$10$ $>$ $75$:$25$ $>$ $50$:$50$.

\subsection{Results}

\begin{figure}
  \centering
      \includegraphics[width=\linewidth]{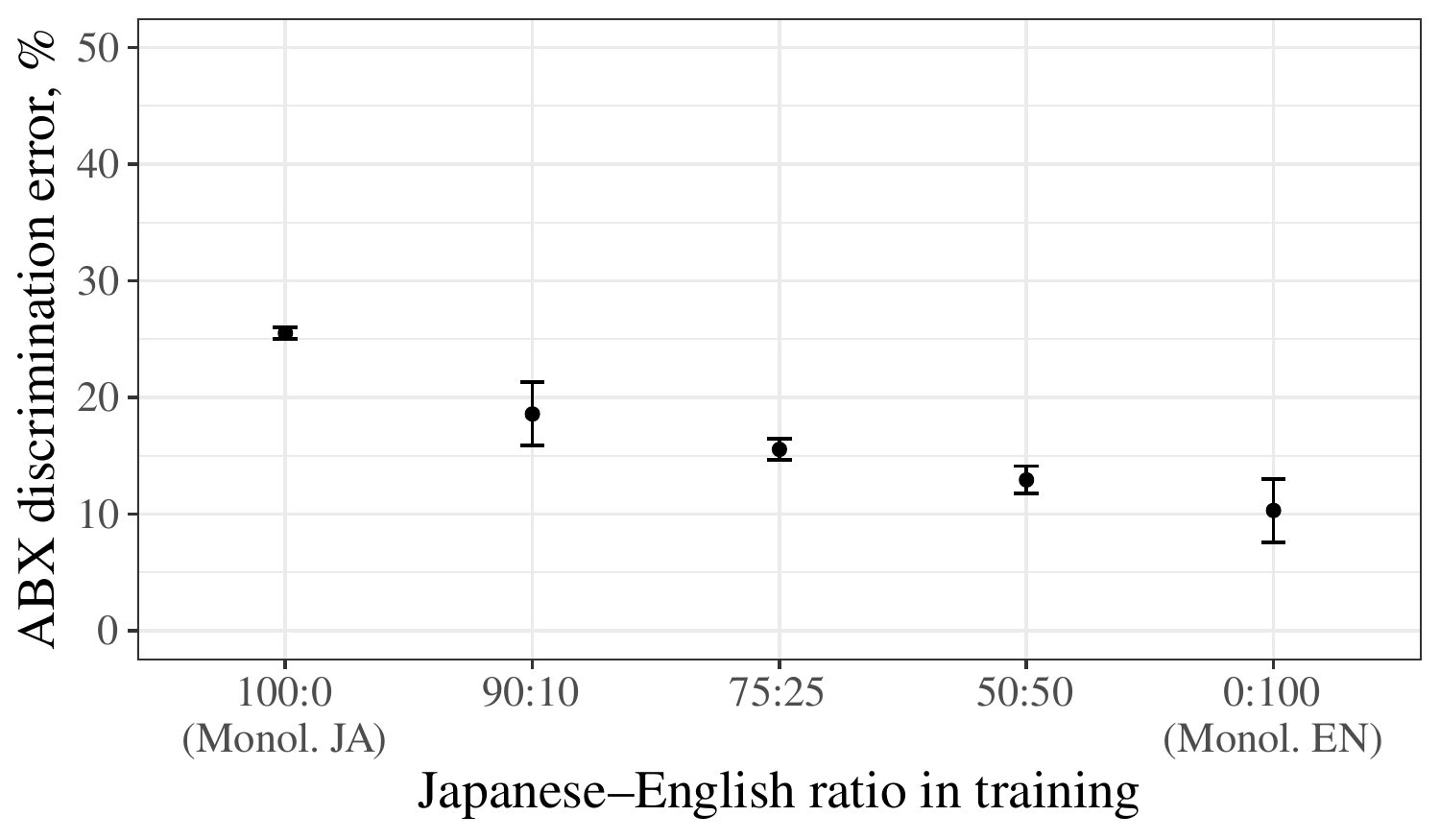}
      \caption{Models' ABX error rates in the word discrimination task with [\textipa{\*r}]--\textipa{[l]} minimal pairs (e.g., \textit{rock--lock}). Error bars show standard error of the mean over two training $\times$ two test corpus samples.}
  \label{fig:jpen}
\end{figure}

It is clear from Figure~\ref{fig:jpen} that the monolingual Japanese model ($100$:$0$) shows higher ABX error rate on the word discrimination task than the monolingual English model ($0$:$100$). This extends the previous result for phone-level [\textipa{\*r}]--\textipa{[l]} discrimination \citep{matusevych2020b} to word level.

Comparing all the models, we observe that the error rate decreases as the relative amount of English exposure increases. Even $10\%$ of English in the training data ($90$:$10$ model) improves the discrimination by $6.9\%$ in absolute error rate compared to the monolingual Japanese model ($25.5\%$ vs.\ $18.6\%$), and more so for the models with higher English exposure.
A mixed-effects regression fitted to the error rates shows that the pairwise differences between most models are statistically significant, except that the $50$:$50$ model shows error rates too close to both its neighbors: $75$:$25$ and $0$:$100$. Despite that, the expected trend is still present. In other words, our model successfully replicates the direction of the main effect in \citet{mackain1981}: the discrimination error rate decreases with higher English exposure.

The result shows that our bilingual model behaves in a predictable way at the word level, ruling out potentially damaging effect that crosslinguistic interactions can have on its lexical representation space. With this knowledge, in the next simulation we proceed with applying our model to a spoken word processing task.

\section{Simulation 3: Spoken word processing}

In this section, we present a case study to show how the model can be used to get a better understanding of spoken word processing. Specifically, we are interested to know if some of the effects reported in the literature can be explained in terms of phonetic learning alone. Difficulties with spoken word processing have been attributed to imprecise phonological encoding of non-native lexical representations \citep{gor2020, cook2016, cook2015, darcy2013, darcy2012}, which results in a spurious activation of similarly sounding competitor words. To give an example from \citet{cook2016}, if the word \textit{parent} is encoded as \textipa{[pEr@(n)t]}, with an optional \textipa{[n]}, it may often be confused with \textit{parrot} \textipa{[pEr@t]}.

We focus on one of the experiments in \citeauthor{cook2016} In a translation judgment task, native Russian speakers (proficient in English) and native English speakers (learning Russian) heard a Russian word (e.g., \textit{moloko} \textipa{[m@\textltilde2"ko]} `milk') and then saw an English word (e.g., \textit{hammer}, which translates into Russian as `molotok' \textipa{[m@\textltilde2"tok]}). The participants had to decide if the English word was a good translation of the Russian one. \citeauthor{cook2016} manipulated the phone edit distance between the true translation and the competitor word: in the example above, the distance between \textipa{[m@\textltilde2"ko]} and \textipa{[m@\textltilde2"tok]} is~$2$. They found that non-native speakers made more mistakes than native speakers, and that increasing the phone edit distance between the target words decreased the size of this effect. They explain the effect by ambiguous (`fuzzy') non-native phonolexical representations. It is unclear, however, whether lexical phonology is necessary to explain the observed effect. To answer this question, we test whether our model with no access to phonology can correctly predict the described effect.

Clearly, our model does not know anything about word meanings and cannot be tested on a translation task. Instead, we use a series of ABX discrimination tasks to test whether the lower performance of non-native speakers can be explained in terms of acoustic embeddings of individual word tokens in the model's representation space. Recall that the model has no access to phonology, and its acoustic embeddings result from phonetic learning alone.

\begin{figure}
  \centering
      \includegraphics[width=\linewidth]{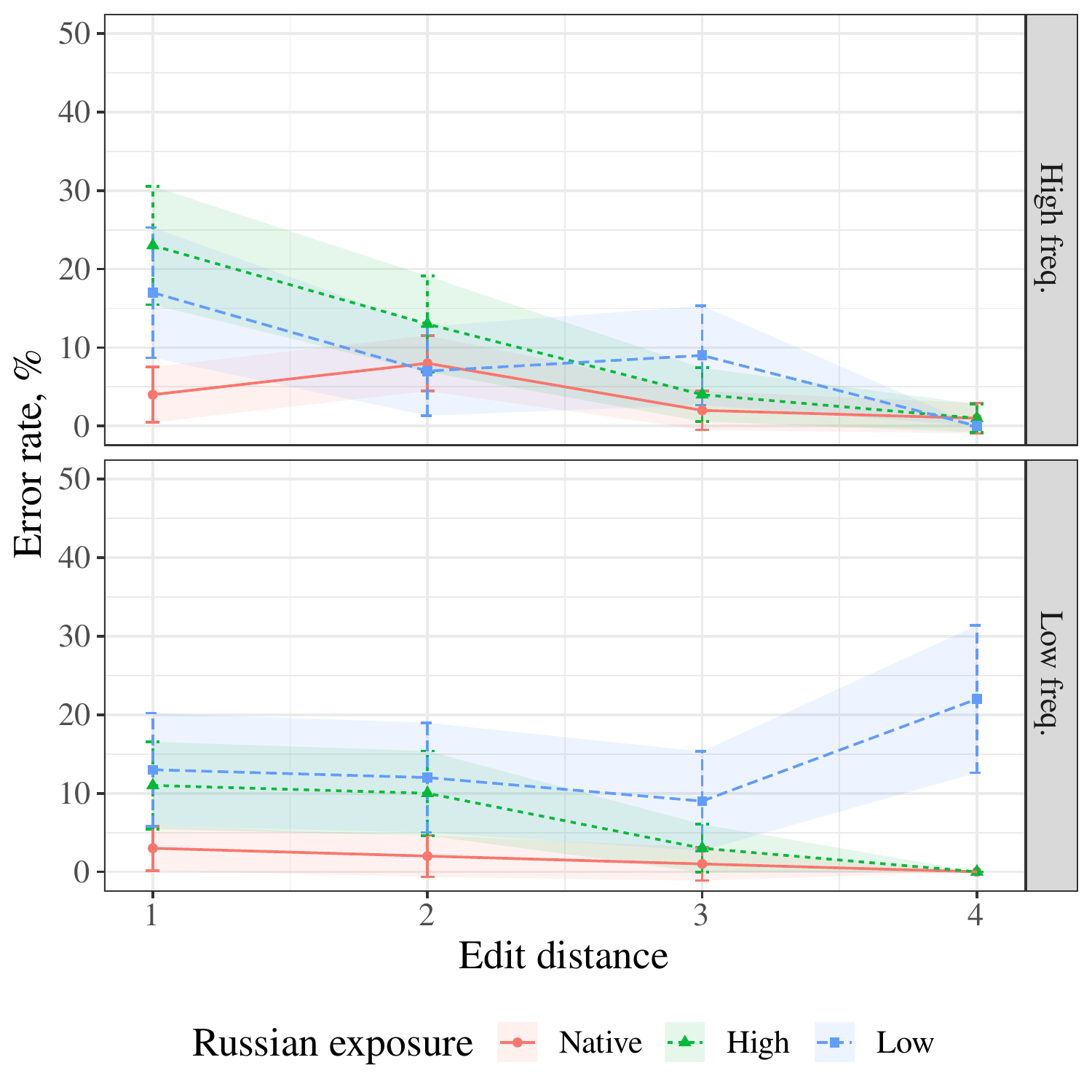}
      \caption{Average error rate of participants in the Russian translation judgment task of \citet{cook2016} depending on their amount of exposure to Russian and the edit distance between the target and the competitor word. Error bars show mean standard error over participants.
      Based on Table~3 in \citet{cook2016}.}
  \label{fig:human}
\end{figure}

Figure~\ref{fig:human} shows the error rates of human participants in \citet{cook2016}. The patterns that we focus on are: (1) lower error rate is associated with higher proficiency (exposure to Russian) and higher edit distance; and (2) the difference between the proficiency groups is the highest when the phone edit distance between the target and the competitor word is low. \citeauthor{cook2016} also looked at the effect of competitor word frequency (hence the two panels in Figure~\ref{fig:human}), but we do not consider this effect here.

\begin{figure*}
 \centering
  \includegraphics[width=0.9\linewidth]{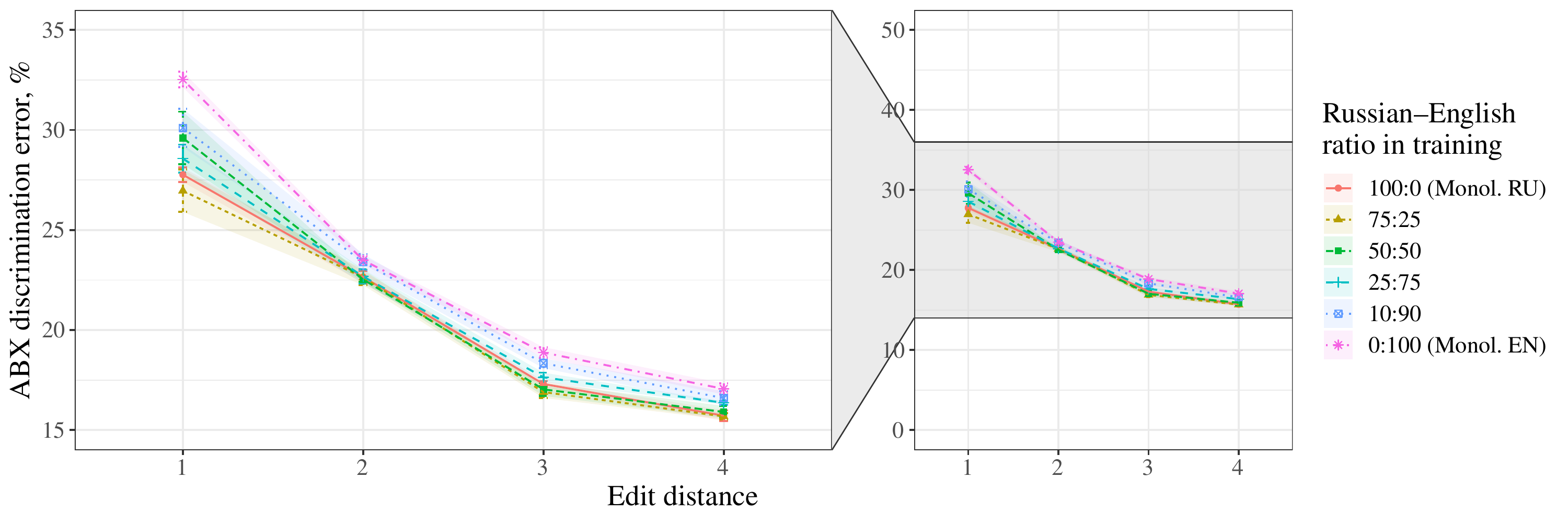}
  \caption{Average ABX discrimination error rate depending on the amount of the model's exposure to Russian and the edit distance between the words in ABX triplets. Error bars show mean standard error over random model initializations.}
  \label{fig:sim3}
\end{figure*}

\subsection{Setup}

We train the model on only Russian or only English data (for reference), and also on the combination of the two in proportion $75$:$25$ (to simulate native Russian speakers with some knowledge of English), $50$:$50$, $25$:$75$, or $10$:$90$ (to simulate native English speakers with variable proficiency in Russian). Each model is trained five times with different random initializations.

For testing, we prepare four ABX discrimination tasks: in each task, the words A and X are of the same type, and the words B and X differ in $1$, $2$, $3$ or $4$ phones (phone edit distance). We could not obtain the list of the original stimuli from \citet{cook2016}, therefore we sample ABX triplets from our test corpus subset. Following the original experiment, we only consider words containing $4$--$10$ phones. Furthermore, we only consider triplets in which all pairwise ratios of the absolute durations of the words are within the factor of $1.1$ (we know from previous work that this model is sensitive to the absolute duration of the test stimuli) and in which B and X are not morphological forms of the same word (such stimuli could not have been used in the original translation judgment task by design). We then sampled $5000$ triplets per task, except for edit distance $1$ we only had $766$ triplets. 

As in the original experiment, we first look at the error rates within each model: the error rates are expected to decrease with greater edit distance. Second, we compare the bilingual models to each other: to match the findings from the human study, models with less exposure to Russian ($50$:$50$, $25$:$75$, $10$:$90$) must show a higher ABX discrimination error than the model with more exposure ($75$:$25$). As a sanity check, we also consider the monolingual Russian ($100$:$0$) and English ($0$:$100$) models.

\subsection{Results}

Figure~\ref{fig:sim3} shows the model's ABX error rates across tasks and training conditions. We first observe that all lines have a negative slope: all the six models show lower error in the tasks with greater edit distance between the words, which is the expected pattern. Note that comparing the absolute values to the results of human participants (Figure~\ref{fig:human}) is not necessarily meaningful because of the task difference: human participants of \citet{cook2016} had to compare an acoustic word in Russian to a translation of an English word they saw, whereas our model directly compared two Russian acoustic words embedded in its representation space.

Second, we see in Figure~\ref{fig:sim3} that the models with less exposure to Russian have higher error rates. This is especially evident for the data points with edit distance $1$, whereas the difference across the models gets smaller with greater edit distances. Again, this is the expected pattern. A mixed-effects regression fitted to the models' error rates suggests that there are significant effects of (1) the amount of Russian language exposure (higher exposure is associated with lower error) and (2) the edit distance (higher edit distance is associated with lower error). A similar pattern is observed when we only consider the bilingual models, for a better analogy with the original experiment: the $75$:$25$ model shows significantly lower error rates than both the $10$:$90$ model and the $25$:$75$ model (but not the $50$:$50$ model), and also the error rates generally get lower as the edit distance increases.

Recall that in each discrimination task, we considered pairs of words with a certain edit distance. Most edit operations involve Russian phone contrasts that are also linguistically meaningful in English (e.g., \textipa{[v]}--\textipa{[s]}). However, there is a small number of Russian phone contrasts that are allophones of the same phoneme in English (e.g., \textipa{[d]}--\textipa{[d\super j]}). If our data included a substantial number of contrasts of this type, the model's higher error rates could be attributed to these difficult phone contrasts. To ensure that was not the case, we looked at the contrasts in our pairs with edit distance $1$ and found that out of $29$ contrasts present in that data, only $1$ (\textipa{[\textltilde]}--\textipa{[l\super j]}) was not phonemic in English, and excluding the corresponding test pairs from the analysis had only minor impact on the absolute error rates, but not on the reported patterns.

To summarize, our model could correctly predict the direction of the two main effects found in a translation judgment task of \citet{cook2016}. This suggests that their result can at least partially be explained in terms of comparing two acoustic instances: the word a participant hears and the translation of a word that (s)he sees. This presents an alternative explanation of the non-native speakers' difficulties with spoken word processing in terms of phonetic perception, which does not involve phonology.

\section{Analyzing the model's representations}

Most studies on bilingual lexical access advocate its non-selective nature: that is, speakers activate words in both languages in parallel, including in spoken word processing \citep[e.g.,][]{weber2004, lagrou2011, shook2012}. Ideally, our model should show a similar pattern and not completely separate the two languages in its representation space. To examine this, we run a language classification task, similar to \citet{kamper2020b}. We are interested whether the model can identify the language of a given word based on its acoustic embedding.

Using the bilingual models from Simulation 3, we embed $5000$ words per language.
We then train a logistic regression classifier on $80\%$ of this data to predict the language of a given word from its acoustic embedding, and test the classifier on the remaining $20\%$ of words. The higher the accuracy of the classifier, the more linearly separable the two languages---specifically, their lexicons---in our model's representations.
The results (Table~\ref{tbl:lid}) show that all models reach accuracy much higher than the $50\%$ chance, although no model reaches $100\%$ accuracy. This means that the lexical representations of words in two languages (acoustic word embeddings) in our bilingual models are not fully linearly separable, indicating a substantial ($13.5$--$19.7\%$) overlap between the two languages. Because some of the representations from the two languages are close to each other in the embedding space, the model may confuse them, similar to the non-selective lexical access in bilingual speakers.

\begin{table}[]
\centering
\begin{tabular}{lcrrr}
\hline
     & \multicolumn{4}{c}{Russian--English ratio}                                                                                 \\ \cline{2-5} 
     & $75$:$25$                  & \multicolumn{1}{c}{$50$:$50$} & \multicolumn{1}{c}{$25$:$75$} & \multicolumn{1}{c}{$10$:$90$} \\ \hline
Mean & \multicolumn{1}{r}{$84.6$} & $86.5$                        & $84.3$                        & $80.3$                        \\
SD   & \multicolumn{1}{r}{$1.0$}  & $0.9$                         & $0.4$                         & $1.6$                         \\ \hline
\end{tabular}
\caption{Accuracy (in \%) of logistic regression classifiers predicting language identity of a given word from its acoustic embedding, averaged over five random initializations of each model.}
\label{tbl:lid}
\end{table}

\section{Discussion}

We started by asking whether some of the difficulties in non-native spoken word processing can be explained at the level of phonetic perception, without involving phonolexical representations. To address this question, we presented a case study (Simulation 3) with a computational model that learns from unsegmented speech data and does not have access to phonology. Our model showed patterns similar to those found by \citet{cook2016} in human speakers. This suggests that their results can be at least partly explained by phonetic learning. While we cannot estimate the relative contribution of the two factors---non-native phonetic perception vs.\ imprecise phonolexical representations---to the behavior of non-native speakers in the experiment of \citeauthor{cook2016}, we argue that both factors need to be considered as possible explanations of the spoken word processing difficulties in non-native speakers. Note, however, that this result does not tell us whether the phonetic or the phonolexical explanation is more parsimonious---a question that should be addressed in the future.

One could interpret our main result differently: that our model, in fact, has succeeded in learning phonological systems from speech data and cannot be considered a purely phonetic model. Indeed, we know that deep neural networks can learn to encode various types of linguistic structure without explicit supervision \citep[e.g.,][]{manning2020, linzen2020}. In particular, speech models can achieve high accuracy in phone discrimination \citep{alishahi2017} and classification \citep{chung2019}, a finding sometimes interpreted as a successful acquisition of phonetic/phonological \textit{categories}. While our model can discriminate at least some phone contrasts, too (Simulation~1), this does not necessarily mean that it learns phonetic categories \citep[see][for a relevant discussion]{schatz2019}. More importantly, what our model does not do is store explicit phonolexical representations in its memory, whereas the (imprecise) storage of word forms is one of the key premises of the phonolexical account explaining non-native speakers' difficulties in spoken word processing \citep{cook2016}. Therefore, we conclude that our results highlight the effects of phonetic perception on non-native word processing.

In Simulation 1 and 2 we showed that our model trained simultaneously on two languages could correctly predict some phone- and word-level discrimination effects in infants and adults \citep{kuhl2003, mackain1981}. This extends previous results on phone discrimination with monolingual model \citep{matusevych2020b} to word discrimination and to bilingual speakers. Also, our analysis of model's representations indicates a substantial overlap between the lexicons of the two languages, mimicking non-selective lexical access in bilingual speakers \citep[e.g.,][]{lagrou2011}. All together, this suggests that the CAE-RNN can be used as a tool to study not only native/non-native phonetic learning, but also native/non-native spoken word processing, including in bilingual speakers.

Our model helps to tease apart the potential impact of phonetic learning from other effects on spoken word processing. At the same time, it is not a cognitive model of the human mental lexicon, for example because it is devoid of semantics. A method to learn acoustic and semantic embeddings in parallel has been proposed in speech engineering \citep{chen2018}, and future research could shed some light on whether this method can be used for studying human mental lexicon.

\section*{Acknowledgments}

This work is based on research supported in part by an ESRC-SBE award ES/R006660/1, a JSMF Scholar Award 220020374, and an NSF award BCS-1734245. We thank the anonymous reviewers, as well as Sameer Bansal, Kate McCurdy, Seraphina Goldfarb-Tarrant and other members of AGORA reading group at the University of Edinburgh for their helpful feedback.

\bibliographystyle{acl_natbib}
\bibliography{references}

\end{document}